# BTREC: BERT-BASED TRAJECTORY RECOMMENDATION FOR PERSONALIZED TOURS

**Ngai Lam Ho, Roy Ka-Wei Lee , Kwan Hui Lim**
Information Systems Technology and Design Pillar
Singapore University of Technology and Design
8 Somapah Road, Singapore 487372
`ngailam_ho@mymail.sutd.edu.sg`, {`roy_lee, kwanhui_lim`}`@sutd.edu.sg`

## ABSTRACT

An essential task for tourists having a pleasant holiday is to have a well-planned itinerary with relevant recommendations, especially when visiting unfamiliar cities. Many tour recommendation tools only take into account a limited number of factors, such as popular Points of Interest (POIs) and routing constraints. Consequently, the solutions they provide may not always align with the individual users of the system. We propose an iterative algorithm in this paper, namely: BTREC (BERT-based Trajectory Recommendation), that extends from the POIBERT embedding algorithm to recommend personalized itineraries on POIs using the BERT framework. Our BTREC algorithm incorporates users' demographic information alongside past POI visits into a modified BERT language model to recommend a *personalized* POI itinerary prediction given a pair of *source* and *destination* POIs. Our recommendation system can create a travel itinerary that maximizes POIs visited, while also taking into account user preferences for categories of POIs and time availability. Our recommendation algorithm is largely inspired by the problem of sentence completion in natural language processing (NLP). Using a dataset of eight cities of different sizes, our experimental results demonstrate that our proposed algorithm is stable and outperforms many other sequence prediction algorithms, measured by recall, precision, and $F_1$-scores.

**Keywords** Recommendation Systems, Neural Networks, Word Embedding, Self-Attention, Transformer

## 1 Introduction

When planning a trip to foreign countries, the typical approach taken by most visitors is to refer to guidebooks/websites for organizing their daily itineraries, or some may employ tour recommendation systems that provide popular points of interest (POIs) based on their popularity[1, 2]. The Transformer architecture has emerged as a highly competitive solution for many NLP tasks, and has also been successfully applied in other domains such as Computer Vision. Unlike some machine learning models such as Long-Short Term Memory and Recurrent Neural Networks that takes in input one at a time, Transformers process the entire input simultaneously and utilizes the *attention mechanism* to model *context* information for each position in the input sequence. This helps to promote increased *parallelism* and enhances overall performance in training and optimization[3]. In this paper, we propose BTREC, a word embedding model using the Transformer architecture to recommend a series of POIs as an itinerary based on historical data with the consideration of the locations, and also traveling time between these POIs. We make the following contributions in this study:

- We propose PPOIBERT, a Transformer-based word embedding model that recommends POIs as an itinerary sequence based on users' historical data, including their POI visit records and travel time between them, while also considering individual user's travel preference.

- We also propose BTREC as a personalized tour recommendation algorithm that extends the PPOIBERT model, to incorporate additional demographic information about travelers into the PPOIBERT model to enhance the accuracy of predictions.

- Our proposed algorithm is evaluated against other sequence prediction methods in our datasets, which covered 9 cities in our experiments. The results of our experiments indicate that our algorithm can predict itineraries reliably with an *average* $\mathcal{F}_1$-score of 63.24% accuracy across all cities.

- Finally, our proposed algorithm has the advantage of adapting to different scenarios (cities/ datasets) without any modification. Furthermore, we observed an increased performance of up to 6.48% in our *Osaka* dataset, as compared to previous implementations measured in average $\mathcal{F}_1$ score (from 56.25% to 62.73%).

The subsequent sections of this paper are structured as follows: Section 2 presents a background on Tour Recommendation and discusses the state-of-the-art in itinerary prediction. Section 3 provides a formal definition of the problem and introduces the notations used in our solution. Section 4 describes our experimental framework and outlines the baseline algorithms used for solution evaluation. In Section 5, we summarize our findings and discuss potential future extensions of this research.

## 2    Preliminaries

### 2.1    Tour Recommendation

Two related problems in tourism-related recommendations are *itinerary planning* and *next location prediction*. Itinerary planning involves scheduling activities to maximize the trip experience within pre-set budgets[2, 4]. Next location prediction identifies the next Pᴏɪ based on others' trajectories. Personalized tour recommendations use check-in data, like photos, to suggest itineraries based on users' interests and preferences. Previous works have focused on recommending popular Pᴏɪs based on queuing time and ratings, using geo-tagged photos to create various tour recommendations[5, 6].

### 2.2    Sequence Prediction

Sequence prediction is a fundamental problem that involves the prediction of the next word in a sequence based on previously observed words[7]. Unlike other prediction algorithms, the order of items in a sequence is crucial to the solution of the problem, making it a valuable technique for time-series forecasting and product recommendation[8]. In the context of tour recommendation, sequence prediction has been adapted by treating Pᴏɪs as words in Nʟᴘ[9]. Existing solutions for Pᴏɪ prediction often employ word-embedding methods such as Word2Vec and FastText to capture Pᴏɪ-to-Pᴏɪ similarity[10, 11, 12]. Other systems use arrays of agents to dynamically explore various solutions and generate optimal itineraries[6]. Moreover, personalized recommendation for Pᴏɪs has been addressed using Pᴏɪ-embedding techniques, providing a refined representation of Pᴏɪs and their categories[13]. These approaches have contributed to more effective tour recommendation systems.

**Bᴇʀᴛ models**    The Transformer model with its effective *self-attention* mechanism is popular and has been widely adopted in Nʟᴘ and computer vision[14]. One of its notable application is the Bidirectional Encoder Representations from Transformers (Bᴇʀᴛ), which has become the state-of-the-art baseline in Nʟᴘ experiments for achieving high accuracy in classification tasks[6, 15]. The training of Bᴇʀᴛ involves the *Masked Language Model* (Mʟᴍ) and *Next Sentence Prediction* (Nsᴘ) algorithms, combined with a loss function. Mʟᴍ trains a model to predict a randomly *masked* words based on surrounding *context*, while Nsᴘ determines whether two sentences appear consecutively in a given text.

Machine Learning algorithms have been proposed to recommend popular Pᴏɪs[13]. These methods use locational data to predict the next Pᴏɪ such that the user is most likely to visit the check-in location[16]. The PᴏɪBᴇʀᴛ model is first proposed by considering the check-ins and duration of users' trajectories as input to the Bᴇʀᴛ language model for training of the Pᴏɪ-prediction task[17, 18]; the algorithm is used to predict itineraries by regarding: 1) users' trajectories as *sentences*, and, 2) travels visit to Pᴏɪs as words into the training of Bᴇʀᴛ model. PᴏɪBᴇʀᴛ then recommends an itinerary by iteratively predicting the next Pᴏɪ (as next word) to visit using the Mʟᴍ prediction model. Durations of visits to thesePᴏɪs are estimated using a statistic model of *Bootstrapping* with a high *confidence level*[17]. However, these recommendations do not take into account the *user's preferences* when selecting a series of Pᴏɪs based on specific interests.

## 3    Problem Formulation and Algorithm

In this section, we introduce the tour recommendation problem and provide a list of the symbols and terms used in Table 1. Given a city as the query with $|P|$ Pᴏɪs, we denote a traveler, $u \in U$, with $k$ check-in records as a sequence of $(poi, time)$ tuples, $S_h = [(p_1^u, t_1^u), (p_2^u, t_2^u) ... (p_k^u, t_k^u)]$, for all $p_i \in$ POIs and $t_i$ denotes as the time marker of the photo posted to LBSN. The problem addressed in this paper is to recommend a personalized sequence of POIs for a traveler who is more likely to visit in a given city, based on a set of historical trajectories. The starting and ending POIs, denoted as $p_1$ and $p_k \in$ POIs, respectively, are also provided in the problem statement.

### 3.1 PPOIBERT - A Refined BERT Model for *P*ersonalized Itinerary Prediction

Previous works have demonstrated that BERT can be utilized as an itinerary prediction model by solving a series of MLM problems[14]. However, this recommendation algorithm only treats users' trajectories as a unified set of *corpus*, without considering how different users (tourists) may prefer to visit different POIs based on their individual tour preferences[17]. To address this limitation, we present PPOIBERT, an innovative approach that incorporates users' information by embedding their *preferences* into the training data of the BERT model. We propose a model with the input using users' information with their past itineraries to improve the accuracy of prediction. This is done by mining users' preferences in deciding subsequent POIs to visit. In the original implementation of BERT, MLM model is trained by randomly masking 15% of words to predicting the *masked* words based on surrounding words or context words[14]. Our proposed PPOIBERT algorithm is to predict the *masked* POI (word), based on the context provided by the context sentence *without masks* (representing POI -IDs and their categories, profile user ID $\bar{u} \in \bar{U}$, and their cities/countries of origin). As shown in Fig.1, we use a similar method for generating *corpus* by translating users' trajectories into *sentences* of user-IDs ($\bar{u}$) and POIs (words)[17]. The generated *corpus* is subsequently trained by the PPOIBERT model for the itinerary prediction task. It is achieved by inserting more demographic information in the *corpus*. The time complexity of PPOIBERT is $\mathcal{O}(N \cdot K^2)$, where $N$ and $K$ denote the total number of POIs and lengths of trajectories, respectively.

**Itinerary prediction in PPOIBERT** Building on the success of POIBERT[17], we present a novel approach to make personalized recommendations by incorporating users' information and their past check-in records into the training model of POIBERT. During the training of our PPOIBERT model, the algorithm takes as input a list of User-IDs and their past trajectories to select subsequent POIs. To achieve this, we enhance the training algorithm by including more users' demographic information into the *corpus*, as illustrated in Fig. 1. Subsequently the PPOIBERT model is then trained with some preset *epochs* and hyper-parameters for predicting itinerary. To make a *personalized* itinerary recommendation, the BTREC algorithm first selects a *similar* user in the training dataset, $\bar{u}$ as a *reference* for making itinerary recommendations. This process of predicting a *personalized* itinerary is to *iteratively* solve an MLM query and insert the predicted POI to a proposed itinerary, outlined in Algorithm 1. The remaining part of recommending personalized itineraries is to *repeatedly* query for the most relevant POI between the source and destination POIs with $\bar{u}$ as a preference model and insert predicted POI to the predicted itinerary in Equation 2.

### 3.2 BTREC - BERT based Personalized Trajectory Recommendation Using Demographic Information

The proposed PPOIBERT algorithm, introduced in Section 3.1, utilizes information from past trajectories based on a *reference* user-ID in the training dataset to make predictions by considering preferences of similar users. BTREC extends the PPOIBERT by fine-tuning the prediction algorithm by considering their demographic information, such as cities and countries, in the training of our embedding model, in addition to the past trajectories of users. This is achieved by modifying the *corpus* and *context sequence* of the PPOIBERT model (as described in Section 3.1), incorporating additional information that may influence the decision-making process for selecting the next POIs to visit. Specifically, each *sentence* in the training data is supplemented with *'word'* representing the user's own city/country after the occurrence of the user-ID, as shown in Fig. 1. The aim is to provide relevant training examples to uncover the *underlying embeddings* between POIs and users from different locations. We then develop *personalized* BERT models for making itinerary predictions based on users' demographic locations and other relevant constraints discussed. We evaluate the accuracy and effectiveness of our proposed algorithm in Section 4.

**Prediction of *reference user*** The initial phase of BTREC involves finding a reference user for the recommendation based on the preference profiles of the training dataset, similar to the K-Nearest Neighbors algorithm[19]. The algorithm iterates through all user-IDs to find the most *similar* to the query specification, which is represented as a $(p_u, p_v)$ pair. This process involves solving a series of simple MLM problems to identify the most similar reference user:

$$\textbf{let } u_r = \operatorname*{ArgMax}_{u \in \bar{U}_{train}} (\textbf{Unmask}("\{\texttt{[CLS]}, u, p_u, \texttt{[MASK]}, u, p_v, \texttt{[SEP]}\}")) \tag{1}$$

$$predict(p_u, p_v) = \operatorname*{ArgMax}_{\forall \bar{u} \in \bar{U}} (\textbf{Unmask}("\texttt{[CLS]}, u_r, p_u, .., \texttt{[MASK]}, .., u_r, p_v, \texttt{[SEP]}")) \tag{2}$$

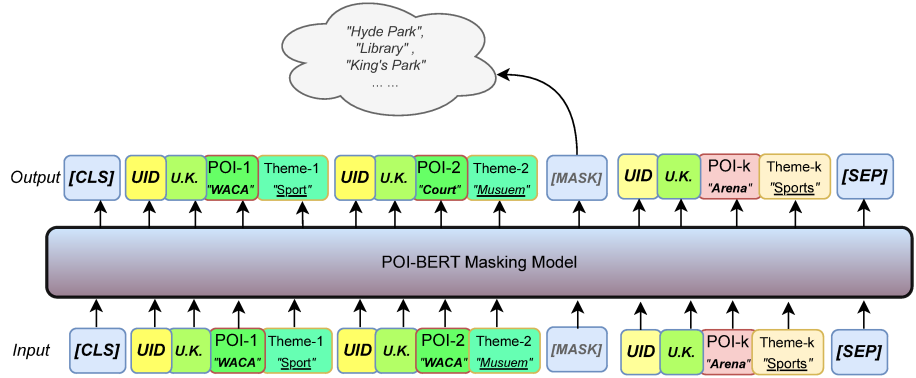

Figure 1: BᴛRᴇᴄ training model

## 4 Experiments and Results

The dataset used in our experimentation comprises photos uploaded in the Flickr platform, encompassing the trajectories of 5,654 users from eight popular cities[20]. The photos also include *meta-data* encompassing details such as the date, time, and GPS locations. By sorting the photos in the dataset based on time and mapping them to the relevant Pᴏɪs using their GPS locations, we can reconstruct the travel trajectories of the users. This process generates sequences of time-sensitive Pᴏɪs that represent the users' trajectories in time. [1]

### 4.1 Datasets

To further evaluate the efficacy of our proposed algorithm on larger data, we incorporated datasets from Melbourne and Vienna[21]; they consist of about 52K photos from 260 Pᴏɪs in these two cities. Our datasets have been divided into three distinct sets: Training, Validation, and Testing datasets. Initially, we sorted all photos according to their Trajectory-IDs based on their *last check-in times* in ascending order. To generate the Training Dataset, we set aside the first 70% of trajectories based on their associated photographs. The subsequent 20% of trajectories were assigned to the *validation* set, while the remaining data was assigned as the *testing dataset*. This method of segregating the data helps to prevent the issue of a having trajectory being present in multiple datasets[17].

**Data:** $p_1^u, p_k^u$: starting/ending Pᴏɪ Ids , $TimeLimit$: time budget of itinerary ;
**Result:** Predicted Pᴏɪ IDs
**begin**
    **let** $q_u \leftarrow$ "[CLS]$, u, p_1^u, c_1,$[MASK]$,$u$,$p_k, c_k,$ [SEP]", $\forall u \in TrainingSet_{user}$ ;
    **let** $u' \leftarrow ArgMax_u(\textbf{Unmask}(q_u))$ ;
    **repeat**
        **for** $\forall j \in \{2..|seq|-1\}$ **do**
            **let** $query_j \leftarrow$ "[CLS]$, u', H_{u'}, p_1^u, c_1, ..., u', H_{u'}, p_{j-1}^u, c_{j-1},$
            [MASK]$, u', H_{u'}, p_j^{u'}, c_j, ..., u', H_{u'}, p_j^{u'}, c_j,$ [SEP]";
        **end**
        **let** $seq \leftarrow \textbf{ArgMax}_j(\textbf{Unmask}(query_j))$ ;
    **until** $TimeLimit < \sum_{poi \in seq} duration(poi)$;
    **return** $seq$ ;
**end**

### 4.2 Baseline Algorithms for Performance Comparison

We compared the performance of our algorithm with the state-of-the-art algorithms of mining sequential patterns. Specifically, we identified the following algorithms for mining sequential patterns for performance comparison:

- Sᴘᴍғ algorithms - the software package consists of several data mining algorithms implemented, such as Cᴘᴛ, Cᴘᴛ+, Tᴅᴀɢ and Markov Chains[22, 23, 24, 25, 26, 27].

---

| | Description |
|---|---|
| $H_u$ | Registered city/country of $u$ |
| $p_j^u$ | Pᴏɪ in Step-$j$ of $u$'s trajectory |
| $p_u$ | source Pᴏɪ |
| $p_v$ | destination Pᴏɪ |
| $S_h$ | Pᴏɪ sequence as a trajectory |
| $S_p$ | Predicted Pᴏɪ sequence |
| $t_j^i$ | Theme label of Pᴏɪ-$p_i$ in step-$j$ of $i$'s trajectory, e.g. 'Museum', 'Park',... etc. |
| $T$ | Total time allocated for the recommended trajectory |
| $\bar{u}_i$ | User-$i$'s preference profile |
| $U$ | set of $\bar{u}_i \in U$, in the training dataset |
| $V^i$ | list of check-ins from user-$i$ sorted by time-stamps as a trajectory, i.e. $V^i = \{v_1^i..v_k^i\}$ |

Table 1: Notation

- SᴜʙSᴇǫ: the algorithm employs compression data structures to efficiently store and manipulate the subsequences as a *"Succinct Wavelet Tree"* data structure[28].

- PᴏɪBᴇʀᴛ: it relies on the general Bᴇʀᴛ model to generate predictions in choosing Pᴏɪs[17]. Additionally, it employs *bootstrapping* to gauge the lengths of Pᴏɪ visits by estimating duration of visits in the Pᴏɪs.

Some baseline algorithms (such as Cᴘᴛ and SᴜʙSᴇǫ) solely predict the subsequent token (as Pᴏɪ,) our sequence prediction task involves *iteratively* predicting additional *tokens* (as Pᴏɪ) until the *pre-set* time limit specified by the user is reached. To compare the effectiveness of our proposed and baselines algorithms, we conducted all experiments under identical conditions outlined in Section 4.3, whereby the algorithms also shared the same datasets for *training*, *validation* and *testing*.

### 4.3 Experiment Methodology and Setup

We performed experiments on eight cities from the Flickr dataset. We considered all trajectories in the dataset as the *ground-truth* dataset for our predictions, and we used the *source/ destination* Pᴏɪs of each trajectory as inputs to our recommendation algorithm. Therefore, we filtered past trajectories with at least 3 Pᴏɪs. To evaluate the performance of our models, we conducted a comparison of the accuracy with different sequence prediction models against various baseline algorithms. The accuracy of our algorithm is evaluated using the *Validation* and *Test* sets as follows: (1) For each trajectory in the dataset, which we refer to as the *history-list*, we considered the first and last Pᴏɪs as the *source* and *destination* Pᴏɪs for the query itinerary. (2) Time limit for the *query* is regarded as the time interval between the first and last photos of each trajectory. (3) Recommend the *intermediate* Pᴏɪs of the trajectory within a specified time frame. To gauge the effectiveness of our models, we compared with various sequence prediction models listed in Section 4.2. The accuracy of these models was assessed using Validation and Test sets. For each trajectory in the dataset, referred to as the *history-list*, we identified the first and last Pᴏɪs as the *source* and *destination* Pᴏɪs for the itinerary prediction query. The time allocated for the query was determined as the time difference between the first and last photos of each trajectory. We evaluated the performance of the proposed BᴛRᴇᴄ prediction algorithm by using the precision ($\mathcal{T_P}$), recall ($\mathcal{T_R}$), and $\mathcal{F}_1$ scores[17].

**Tuning of hyper-parameters**  To find the optimal hyperparameters for our experiments, we trained our models in PPᴏɪBᴇʀᴛ (and BᴛRᴇᴄ) using various *epochs*, ranging from 1 to 100, on the *training* dataset. Next, we employed these trained models to predict itineraries in our *validation* dataset. The model with the highest average $\mathcal{F}_1$ score from the *validation* dataset was selected. Finally, we reported the prediction accuracy using the chosen model to generate recommendations in the *test* dataset. This experiment ensures that we solely rely on the *trained model* to verify its validity in predicting new data.

### 4.4 Experimental Results

We assessed the effectiveness of our proposed algorithm in various cities by comparing the actual Pᴏɪs trajectories (constructed travel histories based on the chronological ordering of photos) as the ground truth dataset values of the itinerary predictions. The accuracy of the predicted itineraries was compared in terms of average $\mathcal{F}_1$ scores in Table 2. The PᴏɪBᴇʀᴛ algorithm achieved 62.32% on average, the proposed BᴛRᴇᴄ algorithm significantly outperforms the baseline algorithms with average $\mathcal{F}_1$ score of 63.55% on our datasets.

Compared to the actual trajectories (*ground-truth* data), both PPᴏɪBᴇʀᴛ and BᴛRᴇᴄ recommend itineraries with high $\mathcal{F}_1$ scores, suggesting a good balance between the *true positives* and *false positives* in the predictions. Our proposed PPᴏɪBᴇʀᴛ algorithm can recommend tour trajectories that are more personalized and *relevant* to users' preferences. Additionally, our proposed BᴛRᴇᴄ algorithm further enhances the prediction of Pᴏɪ itinerary with users' demographic information into the embedding model. Our BᴛRᴇᴄ algorithm predicted a tour itinerary that further outperforms other baseline algorithms with an average $\mathcal{F}_1$ score: 61.45%. Even without hyper-parameter tuning, the BᴛRᴇᴄ algorithm achieves an average $\mathcal{F}_1$ score of 58.05% across all datasets and hyper-parameters, while the PPᴏɪBᴇʀᴛ algorithm achieves an average $\mathcal{F}_1$ score of 56.45% on average.

| Alg. | | Budapest | Delhi | Edinburgh | Glasgow | Melbourne | Osaka | Perth | Toronto | Vienna | *All cities* |
|---|---|---|---|---|---|---|---|---|---|---|---|
| | $\mathcal{R}$ | 64.36 | 82.22 | 68.38 | 71.82 | 24.92 | 58.33 | 61.67 | 76.21 | 61.33 | 66.44 |
| CPT | $\mathcal{F}_1$ | 49.69 | 53.57 | 51.47 | 63.88 | 39.35 | 37.78 | 52.38 | 57.79 | 46.54 | 49.54 |
| | $\mathcal{P}$ | 63.28 | 64.45 | 61.97 | 71.97 | 100.0³ | 55.83 | 81.25 | 63.47 | 59.12 | 63.89 |
| | $\mathcal{R}$ | 64.36 | 66.18 | 73.14 | 72.89 | 24.92 | 52.37 | 66.67 | 74.17 | 59.33 | 66.43 |
| CPT+ | $\mathcal{F}_1$ | 59.63 | 60.38 | 54.72 | 59.91 | 39.35 | 58.22 | *64.59* | 63.10 | 56.45 | 60.20 |
| | $\mathcal{P}$ | 63.28 | 62.56 | 48.09 | 57.04 | 100.0³ | 75.04 | 76.04 | 68.94 | 59.22 | 64.77 |
| | $\mathcal{R}$ | 66.40 | 62.29 | 71.78 | 68.79 | 24.92 | 72.90 | 71.66 | 72.11 | 60.63 | 66.85 |
| DG | $\mathcal{F}_1$ | 57.37 | *69.85* | *62.58* | **64.82** | 39.35 | 63.10 | 57.39 | 63.71 | 57.81 | 60.74 |
| | $\mathcal{P}$ | 57.33 | 75.00 | 61.03 | 72.73 | 100.0³ | 56.25 | 49.45 | 61.55 | 60.23 | 60.43 |
| | $\mathcal{R}$ | 65.15 | 62.29 | 70.35 | 48.57 | 7.28 | 66.43 | 58.33 | 77.90 | 62.23 | 62.71 |
| LZ78 | $\mathcal{F}_1$ | 56.89 | *69.85* | 59.31 | 48.18 | 39.35 | **66.67** | 57.48 | 62.88 | 58.72 | 58.75 |
| | $\mathcal{P}$ | 57.50 | 82.92 | 57.69 | 54.95 | 100.0³ | 68.75 | 62.33 | 56.90 | 62.08 | 61.86 |
| | $\mathcal{R}$ | 63.16 | 100 | 70.61 | 63.64 | 24.92 | 58.33 | 64.17 | 72.11 | 60.84 | 68.92 |
| Markov Chain | $\mathcal{F}_1$ | 56.22 | 62.63 | 56.06 | 64.76 | 39.35 | 51.79 | 63.99 | 63.71 | *59.66* | 59.80 |
| | $\mathcal{P}$ | 57.40 | 47.42 | 51.48 | 65.91 | 100.0³ | 47.50 | 77.50 | 61.55 | 64.30 | 59.39 |
| | $\mathcal{R}$ | 64.32 | 64.32 | 71.73 | 57.12 | 24.92 | 58.33 | 64.17 | 77.31 | 54.56 | 62.87 |
| TDAG | $\mathcal{F}_1$ | 55.57 | 67.59 | 59.09 | 56.29 | 24.92 | 56.94 | 63.99 | 63.40 | 54.56 | 57.90 |
| | $\mathcal{P}$ | 55.57 | 54.92 | 55.84 | 48.19 | 100.0³ | 55.83 | 77.50 | 58.23 | 56.05 | 56.99 |
| | $\mathcal{R}$ | 31.98 | 28.96 | 31.29 | 41.97 | 24.92 | 38.67 | 48.33 | 32.29 | 34.06 | 34.80 |
| SubSeQ | $\mathcal{F}_1$ | 40.33 | 41.67 | 40.97 | 55.04 | 39.35 | 44.38 | 54.05 | 40.18 | 42.88 | 44.06 |
| | $\mathcal{P}$ | 60.80 | 81.25 | 66.14 | 87.12 | 100.0³ | 58.33 | 65.00 | 60.20 | 63.27 | 68.92 |
| | $\mathcal{R}$ | 58.87 | 88.89 | 66.38 | 75.45 | 45.37 | 46.67 | 95.00 | 83.33 | 73.07 | 61.16 |
| PᴏɪBᴇʀᴛ | $\mathcal{F}_1$ | **59.95** | 62.63 | 59.75 | 62.70 | 45.37 | 57.94 | 62.96 | *63.92* | 62.32 | 62.32 |
| | $\mathcal{P}$ | 70.88 | 51.39 | 65.54 | 62.85 | 43.32 | 77.78 | 52.40 | 54.17 | 51.45 | 73.84 |
| | $\mathcal{R}$ | 59.40 | 64.44 | 64.28 | 72.73 | 54.57 | 72.92 | 69.44 | 63.60 | 66.61 | 65.01 |
| BᴛRᴇᴄ | $\mathcal{F}_1$ | *58.69* | **73.89** | **62.83** | *64.81* | **49.58** | 65.58 | **66.07** | **66.13** | **60.86** | **63.55** |
| | $\mathcal{P}$ | 66.73 | 88.80 | 70.69 | 67.07 | 55.50 | 62.50 | 80.00 | 74.34 | 64.44 | 70.10 |

³ these algorithms cannot find new Pᴏɪ, except from the starting and ending Pᴏɪs. Hence, they have precision score: 100%.

Table 2: Average Recall($\mathcal{R}$)/$\mathcal{F}_1$/Precision($\mathcal{P}$) scores of prediction algorithms in Test datasets (%)

## 5 Conclusion

In this paper, we introduce BᴛRᴇᴄ designed to suggest a sequence of Pᴏɪs that enables tourists to plan an optimal schedule while considering factors such as locality, time constraints, and individual preferences. Our approach involves constructing and training a Bᴇʀᴛ-based language model to fine-tune the recommendation system. This process utilizes training, validation, and test datasets to ensure accurate and personalized recommendations. By leveraging the power of Bᴇʀᴛ classification, we aim to provide tourists with a more *refined* and *context-aware* itinerary planning. Additionally, we designed an iterative method to generate Pᴏɪs based on users' interests and demographic information. By analyzing just a pair of source and destination POIs, our iterative algorithm, BTREC, accurately identify users' preferences for selecting subsequent POIs during their tours by analyzing the statistics of uploading (potentially) photos over the frame of their visits to POIs. Extensive experiments conducted on eight cities showed that our proposed algorithm, which considers frequencies of photos, locality of POIs with other users' demographic information, outperforms eight baseline algorithms in terms of average $F_1$ scores. A potential extension of our work involves fine-tuning personalized embeddings for users with *missing* demographic information for more accurate recommendations.

## Acknowledgments

This research is funded in part by the Singapore University of Technology and Design under grant RS-MEFAI-00005-R0201. The computational work was partially performed on resources of the National Super-Computing Centre and the Social AI Studio, SUTD.